\documentclass[11pt,a4paper]{article}
\usepackage{mtsummit2019}
\usepackage{times}
\usepackage{url}
\usepackage{latexsym}
\usepackage[small,bf]{caption} 
\usepackage[utf8]{inputenc} 
\usepackage[T1]{fontenc}    

\makeatletter
\newcommand{\@BIBLABEL}{\@emptybiblabel}
\newcommand{\@emptybiblabel}[1]{}
\makeatother
\usepackage{booktabs}       
\usepackage{amsfonts}       
\usepackage{nicefrac}       
\usepackage{microtype}      
\usepackage{multirow}
\usepackage{caption}
\usepackage{amsmath,amssymb}
\usepackage{dsfont}			
\usepackage{cases}
\usepackage{color}
\usepackage[table,xcdraw]{xcolor}
\usepackage{tikz}
\usepackage{pgfplots, pgfplotstable}
\usepgfplotslibrary{statistics}
\usepackage{filecontents}
\usetikzlibrary{shapes,arrows,calc}
\usepackage{subfig}
\usepackage{caption}
\usepackage{textcomp}
\usepackage{enumitem}
\usepackage{import}
\usepackage{CJKutf8}
\usepackage[toc,page]{appendix}
\usepackage[textsize=large]{todonotes}
\usepackage[normalem]{ulem}
\useunder{\uline}{\ul}{}

\title{A Call for Prudent Choice of Subword Merge Operations\\ in Neural Machine Translation}

\author{Shuoyang Ding$\dagger$\quad Adithya Renduchintala$\dagger$\quad Kevin Duh$\dagger\ddagger$\\
$\dagger$ Center for Language and Speech Processing\\
$\ddagger$ Human Language Technology Center of Excellence\\
Johns Hopkins University\\
{\tt \{dings, adi.r\}@jhu.edu\quad kevinduh@cs.jhu.edu}}

\date{}

\begin{document}
\maketitle
\begin{abstract}
Most neural machine translation systems are built upon subword units extracted by methods such as Byte-Pair Encoding (BPE) or wordpiece.
However, the choice of \textit{number of merge operations} is generally made by following existing recipes.
In this paper, we conduct a systematic exploration on different numbers of BPE merge operations to understand how it interacts with the model architecture, the strategy to build vocabularies and the language pair.
Our exploration could provide guidance for selecting proper BPE configurations in the future.
Most prominently: we show that for LSTM-based architectures, it is necessary to experiment with a wide range of different BPE operations as there is no typical optimal BPE configuration, whereas for Transformer architectures, smaller BPE size tends to be a typically optimal choice.
We urge the community to make prudent choices with subword merge operations, as our experiments indicate that a sub-optimal BPE configuration alone could easily reduce the system performance by 3--4 BLEU points.
\end{abstract}

\section{Introduction}
While achieving state-of-the-art results, it is a common constraint that Neural Machine Translation (NMT)~\cite{dblp:conf/nips/sutskevervl14,dblp:journals/corr/bahdanaucb14,dblp:conf/emnlp/luongpm15,DBLP:conf/nips/VaswaniSPUJGKP17} systems are only capable of generating a closed set of symbols.
Systems with large vocabulary sizes are too hard to fit onto GPU for training, as the word embedding is generally the most parameter-dense component in the NMT architecture.
For that reason, subword methods, such as Byte-Pair Encoding (BPE)~\cite{DBLP:conf/acl/SennrichHB16a}, are very widely used for building NMT systems.
The general idea of these methods is to exploit the pre-defined vocabulary space optimally by performing a minimum amount of word segmentations in the training set.

However, very few existing literature carefully examines what is the best practice regarding application of subword methods. 
As hyper-parameter search is expensive, there is a tendency to simply use existing recipes. 
This is especially true for the \textit{number of merge operations} when people are using BPE, although this configuration is closely correlated with the granularity of the segmentation on the training corpus, thus having direct influence on the final system performance.
Prior to this work, \newcite{DBLP:conf/aclnmt/DenkowskiN17} recommended 32k BPE merge operation in their work on trustable baselines for NMT, while \newcite{DBLP:conf/emnlp/CherryFBFM18} contradicted their study by showing that character-based models outperform 32k BPE.
Both of these studies are based on the LSTM-based architectures~\cite{dblp:conf/nips/sutskevervl14,dblp:journals/corr/bahdanaucb14,dblp:conf/emnlp/luongpm15}.
To the best of our knowledge, there is no work that looks into the same problem for the Transformer architecture extensively.\footnote{For reference, the original Transformer paper by \newcite{DBLP:conf/nips/VaswaniSPUJGKP17} used BPE merge operations that resulted in 37k joint vocabulary size.}

In this paper, we aim to provide guidance for this hyper-parameter choice by examining the interaction between MT system performance with the choice of BPE merge operations under the \textit{low-resource setting}. 
We conjecture that lower resource systems will be more prone to the performance variance introduced by this choice, and the effect might vary with the choice of model architectures and languages.
To verify this, we conduct experiments with 5 different architecture setup on 4 language pairs of IWSLT 2016 dataset.
In general, we discover that there is no typical optimal choice of merge operations for LSTM-based architectures, but for Transformer architectures, the optimal choice lays between 0--4k, and systems using the traditional 32k merge operations could lose as much as 4 points in BLEU score compared to the optimal choice.

\section{Related Work}

\begin{figure}[]
\centering
\scalebox{0.9}{\import{}{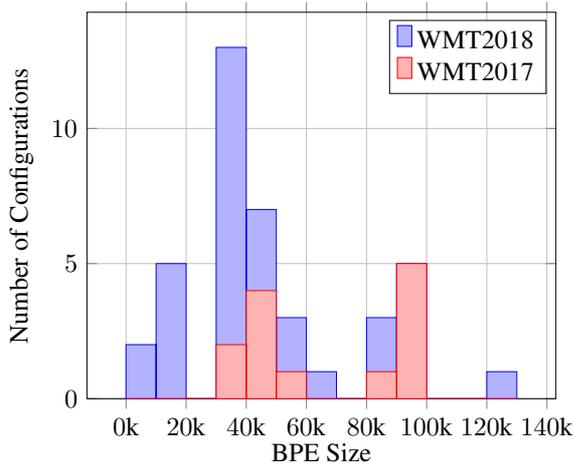}}
\caption{Histogram of BPE merge operations used for in WMT papers from 2017-2018.
\label{fig:pie}}
\end{figure}

Currently, the most common subword methods are BPE~\cite{DBLP:conf/acl/SennrichHB16a}, wordpiece~\cite{DBLP:journals/corr/WuSCLNMKCGMKSJL16} and subword regularization~\cite{DBLP:conf/acl/Kudo18}. 
Subword regularization introduces Bayesian sampling method to incorporate more segmentation variety into the training corpus, thus improving the systems' ability to handle segmentation ambiguity.
Yet, the effect of such method is not very thoroughly tested.
In this work we will focus on the BPE/wordpiece method. 
Because the two methods are very similar, throughout the rest of the paper, we will refer to the BPE/wordpiece method as \textit{BPE method} unless otherwise specified.

To the best of our knowledge, no prior work systematically reports findings for a wide range of systems that cover different architectures and both directions of translation for multiple language pairs.
While some work has conducted experiments with different BPE settings, they are generally very limited in the range of configurations explored.
For example, \newcite{DBLP:conf/acl/SennrichHB16a}, the original paper that proposed the BPE method, compared the system performance when using 60k separate BPE and 90k joint BPE.
They found 90k to work better and used that for their subsequent winning WMT 2017 new translation shared task submission \cite{DBLP:conf/wmt/SennrichBCGHHBW17}.
\newcite{DBLP:journals/corr/WuSCLNMKCGMKSJL16}, on the other hand, found 8k--32k merge operations achieving optimal BLEU score performance for the wordpiece method.
\newcite{DBLP:conf/aclnmt/DenkowskiN17} explored several hyperparameter settings, including number of BPE merge operations, to establish strong baseline for NMT on LSTM-based architectures.
While \newcite{DBLP:conf/aclnmt/DenkowskiN17} showed that BPE models are clearly better than word-level models, their experiments on 16k and 32k BPE configuration did not show much difference.
They therefore recommended ``32K as a generally effective vocabulary size and 16K as a contrastive condition when building systems on less than 1 million parallel sentences''.
However, while studying deep character-based LSTM-based translation models, \newcite{DBLP:conf/emnlp/CherryFBFM18} also ran experiments for BPE configurations between 0--32k, and found that the system performance deteriorates with the increasing number of BPE merge operations.
Recently, \newcite{DBLP:journals/corr/abs-1809-02223} also showed that it is important to tune the number of BPE merge operations and found no typical optimal BPE configuration for their LSTM-based architecture while sweeping over several language pairs in the low-resource setting.
It should be noticed that the results from the above studies actually contradict with each other, and there is still no clear consensus as to what is the best practice for BPE application.
Moreover, all the work surveyed above was done with LSTM-based architectures. To this day, we are not aware of any work that explored the interaction of BPE with the Transformer architecture.

To give the readers a better landscape of the current practice, we gather all 44 papers that have been accepted by the research track of Conference of Machine Translation (WMT) through 2017 and 2018. 
We count different configurations used in a single paper as separate data points. Hence, after removing 8 papers for which BPE is irrelevant, we still manage to obtain 42 data points, shown in Figure~\ref{fig:pie}.
It first comes to our attention that 30k--40k is the most popular range for the number of BPE merge operations.
This is mostly driven by the popularity of two configurations: 30k and 32k.
80k--100k is also pretty popular, which is largely due to configurations 89.5k and 90k.
Upon closer examination, we realized that most papers that used 90k were following the configuration in \newcite{DBLP:conf/wmt/SennrichBCGHHBW17}, the winning NMT system in the WMT 2017 news translation shared task, but this setup somehow became less popular in 2018.
On the other hand, although we are unable to confirm a clear trend-setter, 30k--50k always seems to be a common choice. 
Moreover, although smaller BPE size got more popular among configurations in 2018, none of the work published in WMT has ever explored BPE size lower than 6k.
All of the above observations support our initial claim that we as a community have not yet systematically investigated the entire range of BPE merge operations used in our experiments.

\section{Analysis Setup} 
Our goal is to compare the impact of different numbers of BPE merge operations on multiple language pairs and multiple NMT architectures. We experiment with the following BPE merge operation setup: 0 (character-level), 0.5k, 1k, 2k, 4k, 8k, 16k, and 32k, on both translation directions of 4 language pairs and 5 architectures. Additionally, we include 6 more language pairs (with 2 architectures) to study the interaction between linguistic attributes and BPE merge operations.

\subsection{Dataset}
Our experiments are conducted with the all the data from IWSLT 2016 shared task, covering translation of English (en) from and into Arabic (ar), Czech (cs), French (fr) and German (de).
As this dataset contains multiple dev and test sets, we concatenate all the dev sets into a single dev set and do the same for the test set as well.
To increase language coverage, we also conduct extra experiments with 6 more language pairs from the TED corpus~\cite{DBLP:conf/naacl/QiSFPN18}.
We use Brazilian Portuguese (pt), Hebrew (he), Russian (ru), Turkish (tr), Polish (pl) and Hungarian (hu)  as our extra languages, paired with English.
All the data are tokenized and truecased using the accompanying script from Moses decoder~\cite{DBLP:conf/acl/KoehnHBCFBCSMZDBCH07} before training and applying BPE models.\footnote{Data processing scripts available at \texttt{https://github} \texttt{.com/shuoyangd/prudent-bpe}.}

We use subword-nmt\footnote{\tt\scriptsize https://pypi.org/project/subword-nmt/0.3.5/} to train and apply BPE to our data. Unless otherwise specified, all of our BPE models are trained on the concatenation of the source and target training corpus, i.e. the \textit{joint BPE} scheme in \newcite{DBLP:conf/acl/SennrichHB16a}. We use SacreBLEU~\cite{post-2018-call} to compute BLEU score.\footnote{SacreBLEU signature:\texttt{BLEU+case.mixed+numrefs.1+} \texttt{smooth.exp+tok.13a+version.1.2.12.}}
\subsection{Architecture}
\begin{table*}[]
\centering
\scalebox{0.9}{\begin{tabular}{@{}llllllll@{}}
\toprule
\textbf{} & \textbf{bi-dir} & \textbf{$d_{enc}$} & \textbf{$d_{dec}$} & \textbf{$d_{emb}$} & \textbf{$l$} & \textbf{$N_h$} & \textbf{$N_p$} \\ \midrule
\textbf{shallow-transformer} & N/A & 512 & 512 & 512 & 2 & 4 & 18.8M \\
\textbf{deep-transformer} & N/A & 512 & 512 & 512 & 6 & 4 & 39.8M \\
\textbf{tiny-lstm} & no & 256 & 256 & 256 & 1 & 1 & 5.6M \\
\textbf{shallow-lstm} & yes & 384 & 384 & 384 & 2 & 1 & 16.4M \\
\textbf{deep-lstm} & yes & 384 & 384 & 384 & 6 & 1 & 35.3M \\ \bottomrule
\end{tabular}}
\caption{Information of the 5 architectures used for analysis. \textbf{bi-dir} is a boolean representing whether the encoder is bi-directional. $d_{enc}$, $d_{dec}$ and $d_{emb}$ are dimension of encoder, decoder and source/target word embedding, respectively. $l$ is the number of encoder/decoder layers. $N_h$ is the number of attention heads, while $N_p$ is the number of parameters of the model at 8k BPE merge operations.\label{tab:param-no}}
\end{table*}

We build our NMT system with fairseq~\cite{ott2019fairseq}.
We use two pre-configured architectures in fairseq for our study, namely \texttt{lstm-wiseman-iwslt-de-en} (referred to as \texttt{tiny-lstm}) and \texttt{trans-} \texttt{former-iwslt-de-en} (referred to as \texttt{deep-} \texttt{transformer}), which are the model architecture tuned for their benchmark system trained on IWSLT 2014 German-English data.
However, we find (as can be seen from Table~\ref{tab:param-no}) that the number of parameters in \texttt{lstm-tiny} is a magnitude lower than \texttt{deep-transformer} mainly due to the fact that the former has a single-layer uni-directional encoder and a single-layer decoder, while the later has $6$ encoder and decoder layers.
For a fairer comparison we include a \texttt{deep-lstm} architecture with $6$ encoder and decoder layers which roughly matches the number of parameters in \texttt{deep-transformer}.
To study the effect of BPE on relatively smaller architectures, we also include  \texttt{shallow-transformer}  and  \texttt{shallow-lstm} architectures, both with $2$ encoder and decoder layers. 
The \texttt{shallow-lstm} also use bidirectional LSTM layers in the encoder.
These two architectures also roughly match each other in terms of number of parameters.
With these $5$ architectures, we believe we have covered a wide range of common choices in NMT architectures, especially in low-resource settings.
We use Adam optimizer \cite{kingma2014adam} for all the experiments we run.
For Transformer experiments, we use the learning rate scheduling settings in \newcite{DBLP:conf/nips/VaswaniSPUJGKP17}, including the inverse square root learning rate scheduler, 4000 warmup updates and initial warmup learning rate of $1\times 10^{-7}$.
For most LSTM experiments, we just use learning rate 0.001 from the start and reduce the learning rate by half every time the loss function fails to improve on the development set.
However, we find that for \texttt{deep-lstm} architecture, such learning rate schedule tends to be unstable, which is very similar to training Transformer without the warmup learning rate schedule.
Applying the same warmup schedule as Transformer experiments works for most \texttt{deep-lstm} architecture except for de-en experiments as BPE size 16k and 32k, for which we have to apply 8000 warmup updates. 
Per the experiment setting in \newcite{DBLP:conf/nips/VaswaniSPUJGKP17}, we also apply label smoothing with $\varepsilon_{ls} = 0.1$ for all of our Transformer experiments.


\section{Analysis}
\definecolor{applegreen}{rgb}{0.55, 0.71, 0.0}
\definecolor{bananamania}{rgb}{0.98, 0.91, 0.71}

\colorlet{r1}{applegreen}
\colorlet{r2}{bananamania!14!applegreen}
\colorlet{r3}{bananamania!29!applegreen}
\colorlet{r4}{bananamania!43!applegreen}
\colorlet{r5}{bananamania!58!applegreen}
\colorlet{r6}{bananamania!72!applegreen}
\colorlet{r7}{bananamania!86!applegreen}
\colorlet{r8}{bananamania}

\let\cc\cellcolor
\subsection{Analysis 1: Architectures}

\begin{table*}[]
\centering
\scalebox{0.90}{\begin{tabular}{@{}llrrrrrrrrr@{}}
\toprule
 &  & \multicolumn{1}{l}{\textbf{0}} & \multicolumn{1}{l}{\textbf{0.5k}} & \multicolumn{1}{l}{\textbf{1k}} & \multicolumn{1}{l}{\textbf{2k}} & \multicolumn{1}{l}{\textbf{4k}} & \multicolumn{1}{l}{\textbf{8k}} & \multicolumn{1}{l}{\textbf{16k}} & \multicolumn{1}{l}{\textbf{32k}} & \multicolumn{1}{l}{$\delta$} \\ \midrule
\multirow{8}{*}{\textbf{\begin{tabular}[c]{@{}l@{}}deep-\\ transformer\end{tabular}}} & \textbf{ar-en} & 30.3\cc{r5} & 30.8\cc{r1} & 30.6\cc{r2} & 30.5\cc{r3} & 30.4\cc{r4} & 29.8\cc{r6} & 28\cc{r7} & 27.5\cc{r8} & 3.3 \\
 & \textbf{cs-en} & 24.6\cc{r1} & 23.3\cc{r2} & 23.0\cc{r3} & 22.7\cc{r4} & 21.2\cc{r6} & 22.6\cc{r5} & 20.6\cc{r8} & 21.0\cc{r7} & 4.0 \\
 & \textbf{de-en} & 28.1\cc{r3} & 28.6\cc{r1} & 28.0\cc{r4} & 28.4\cc{r2} & 27.7\cc{r5} & 27.5\cc{r6} & 26.7\cc{r7} & 25.2\cc{r8} & 3.4 \\
 & \textbf{fr-en} & 28.8\cc{r4} & 29.8\cc{r1} & 29.6\cc{r2} & 29.3\cc{r3} & 28.7\cc{r5} & 28.5\cc{r6} & 27.5\cc{r7} & 26.6\cc{r8} & 3.2 \\ \cmidrule(l){2-11}
 & \textbf{en-ar} & 12.6\cc{r2} & 13.0\cc{r1} & 12.1\cc{r4} & 12.3\cc{r3} & 11.8\cc{r5} & 11.3\cc{r6} & 10.7\cc{r7} & 10.6\cc{r8} & 2.4 \\
 & \textbf{en-cs} & 17.3\cc{r1} & 17.1\cc{r2} & 16.7\cc{r3} & 16.4\cc{r4} & 16.1\cc{r5} & 15.6\cc{r6} & 14.7\cc{r7} & 13.8\cc{r8} & 3.5 \\
 & \textbf{en-de} & 26.1\cc{r4} & 27.4\cc{r1} & 27.4\cc{r1} & 26.1\cc{r4} & 26.3\cc{r3} & 26.1\cc{r4} & 25.8\cc{r7} & 23.9\cc{r8} & 3.5 \\
 & \textbf{en-fr} & 25.2\cc{r5} & 25.6\cc{r1} & 25.3\cc{r3} & 25.5\cc{r2} & 25.3\cc{r3} & 24.7\cc{r6} & 24.1\cc{r7} & 22.8\cc{r8} & 2.8 \\ \midrule
\multirow{8}{*}{\textbf{\begin{tabular}[c]{@{}l@{}}shallow-\\ transformer\end{tabular}}} & \textbf{ar-en} & 26.4\cc{r6} & 27.9\cc{r4} & 28.7\cc{r1} & 28.5\cc{r3} & 28.6\cc{r2} & 27.7\cc{r5} & 26.2\cc{r7} & 25.5\cc{r8} & 3.2 \\
 & \textbf{cs-en} & 22.4\cc{r2} & 22.6\cc{r1} & 22.3\cc{r3} & 21.8\cc{r4} & 21.7\cc{r5} & 21.1\cc{r6} & 21.1\cc{r6} & 20.1\cc{r8} & 2.5 \\
 & \textbf{de-en} & 25.5\cc{r6} & 27.4\cc{r1} & 27.1\cc{r3} & 27.3\cc{r2} & 27.1\cc{r3} & 25.9\cc{r5} & 24.6\cc{r7} & 23.7\cc{r8} & 3.7 \\
 & \textbf{fr-en} & 26.3\cc{r6} & 28.0\cc{r2} & 28.9\cc{r1} & 28.0\cc{r2} & 28.0\cc{r2} & 27.4\cc{r5} & 26.1\cc{r7} & 26.1\cc{r7} & 2.7 \\ \cmidrule(l){2-11} 
 & \textbf{en-ar} & 11.7\cc{r1} & 11.2\cc{r4} & 11.5\cc{r2} & 11.0\cc{r5} & 11.3\cc{r3} & 10.5\cc{r6} & 9.5\cc{r7} & 9.0\cc{r8} & 2.7 \\
 & \textbf{en-cs} & 16.4\cc{r2} & 16.7\cc{r1} & 16.0\cc{r4} & 16.2\cc{r3} & 14.4\cc{r5} & 14.2\cc{r6} & 13.9\cc{r7} & 13.9\cc{r8} & 2.8 \\
 & \textbf{en-de} & 23.8\cc{r7} & 25.7\cc{r1} & 25.4\cc{r2} & 25.3\cc{r3} & 25.2\cc{r4} & 24.3\cc{r5} & 24.1\cc{r6} & 22.1\cc{r8} & 3.6 \\
 & \textbf{en-fr} & 23.5\cc{r6} & 24.7\cc{r2} & 25.1\cc{r1} & 24.6\cc{r3} & 24.5\cc{r4} & 23.8\cc{r5} & 22.7\cc{r7} & 22.1\cc{r8} & 3.0 \\ \bottomrule
\end{tabular}}
\caption{BLEU score for Transformer architectures with multiple BPE configurations. Each score is color-coded by its rank among scores from different BPE configurations in the same row. $\delta$ is the difference between the best and worst BLEU score of each row.\label{tab:transformer}}
\end{table*}

Table~\ref{tab:transformer} shows the BLEU score for Transformer systems with BPE merge operations ranging from 0 to 32k.
The Transformer experiments show a clear trend; large BPE settings of $16$k-$32$k are \emph{not} optimal for low-resource settings.
We see that regardless of the direction of translation, the best BLEU score for Transformer-based architectures are somewhere in the $0$-$1$k range.
Although there is not much drop for $2$k-$4$k, there is generally a drastic performance drop as the number of BPE merge operation is increased beyond $8$k.
It should also be noted that the difference between the best and the worst performance is around 3 BLEU points (refer to the $\delta$ column in Table~\ref{tab:transformer}), larger than the improvements claimed in many machine translation papers.


\begin{table*}[]
\centering
\scalebox{0.90}{\begin{tabular}{@{}llrrrrrrrrr@{}}
\toprule
\textbf{} & \textbf{} & \multicolumn{1}{l}{\textbf{0}} & \multicolumn{1}{l}{\textbf{0.5k}} & \multicolumn{1}{l}{\textbf{1k}} & \multicolumn{1}{l}{\textbf{2k}} & \multicolumn{1}{l}{\textbf{4k}} & \multicolumn{1}{l}{\textbf{8k}} & \multicolumn{1}{l}{\textbf{16k}} & \multicolumn{1}{l}{\textbf{32k}} & \multicolumn{1}{l}{$\delta$} \\ \midrule
\multirow{8}{*}{\textbf{\begin{tabular}[c]{@{}l@{}}tiny-\\ lstm\end{tabular}}} & \textbf{ar-en} & 20.6\cc{r8} & 22.1\cc{r7} & 22.4\cc{r6} & 23.0\cc{r5} & 24.1\cc{r3} & 24.2\cc{r1} & 24.2\cc{r1} & 24.0\cc{r4} & 3.6 \\
 & \textbf{cs-en} & 17.8\cc{r8} & 19.1\cc{r4} & 18.8\cc{r7} & 19.0\cc{r6} & 19.2\cc{r3} & 19.5\cc{r2} & 20.7\cc{r1} & 19.1\cc{r4} & 2.9 \\
 & \textbf{de-en} & 21.1\cc{r8} & 22.5\cc{r7} & 23.2\cc{r2} & 23.1\cc{r3} & 23.1\cc{r3} & 23.1\cc{r3} & 23.6\cc{r1} & 23.0\cc{r6} & 2.5 \\
 & \textbf{fr-en} & 21.8\cc{r8} & 25.3\cc{r2} & 25.3\cc{r2} & 25.4\cc{r1} & 25.1\cc{r5} & 25.3\cc{r2} & 25.1\cc{r5} & 24.7\cc{r7} & 3.6 \\ \cmidrule(l){2-11} 
 & \textbf{en-ar} & 8.5\cc{r8} & 8.7\cc{r6} & 9.3\cc{r1} & 8.8\cc{r2} & 8.8\cc{r2} & 8.6\cc{r7} & 8.8\cc{r2} & 8.8\cc{r2} & 0.8 \\
 & \textbf{en-cs} & 11.5\cc{r8} & 12.3\cc{r7} & 13.7\cc{r3} & 13.2\cc{r4} & 13.0\cc{r6} & 14.1\cc{r2} & 14.4\cc{r1} & 13.2\cc{r4} & 2.9 \\
 & \textbf{en-de} & 18.2\cc{r8} & 20.8\cc{r7} & 21.4\cc{r4} & 21.1\cc{r5} & 21.9\cc{r1} & 21.6\cc{r2} & 21.0\cc{r6} & 21.6\cc{r2} & 3.7 \\
 & \textbf{en-fr} & 19.9\cc{r8} & 20.4\cc{r7} & 20.7\cc{r6} & 21.8\cc{r1} & 21.3\cc{r2} & 21.0\cc{r5} & 21.3\cc{r2} & 21.3\cc{r2} & 1.7 \\ \midrule
\multirow{8}{*}{\textbf{\begin{tabular}[c]{@{}l@{}}shallow-\\ lstm\end{tabular}}} & \textbf{ar-en} & 27.5\cc{r2} & 27.2\cc{r5} & 27.1\cc{r6} & 27.6\cc{r1} & 27.4\cc{r4} & 26.7\cc{r7} & 27.5\cc{r2} & 26.3\cc{r8} & 1.3 \\
 & \textbf{cs-en} & 22.2\cc{r5} & 22.2\cc{r5} & 22.2\cc{r5} & 22.9\cc{r2} & 22.7\cc{r4} & 23.0\cc{r1} & 22.8\cc{r3} & 21.6\cc{r8} & 1.4 \\
 & \textbf{de-en} & 25.7\cc{r8} & 25.9\cc{r7} & 26.0\cc{r5} & 25.9\cc{r6} & 26.4\cc{r2} & 26.3\cc{r3} & 26.1\cc{r4} & 26.5\cc{r1} & 0.8 \\
 & \textbf{fr-en} & 27.6\cc{r7} & 26.7\cc{r8} & 27.7\cc{r5} & 28.4\cc{r2} & 27.9\cc{r3} & 27.7\cc{r4} & 28.5\cc{r1} & 27.5\cc{r6} & 1.8 \\ \cmidrule(l){2-11} 
 & \textbf{en-ar} & 11.0\cc{r1} & 11.0\cc{r1} & 10.7\cc{r3} & 10.4\cc{r6} & 10.6\cc{r4} & 10.6\cc{r4} & 10.4\cc{r6} & 10.1\cc{r8} & 0.9 \\
 & \textbf{en-cs} & 16.1\cc{r1} & 15.7\cc{r5} & 15.8\cc{r2} & 15.3\cc{r8} & 15.8\cc{r2} & 15.5\cc{r7} & 15.8\cc{r2} & 15.6\cc{r6} & 0.8 \\
 & \textbf{en-de} & 24.9\cc{r6} & 25.1\cc{r4} & 23.9\cc{r8} & 24.2\cc{r7} & 25.4\cc{r2} & 25.2\cc{r3} & 25.5\cc{r1} & 25.0\cc{r5} & 1.6 \\
 & \textbf{en-fr} & 24.3\cc{r1} & 23.8\cc{r5} & 23.7\cc{r6} & 24.2\cc{r2} & 23.5\cc{r7} & 24.1\cc{r3} & 23.9\cc{r4} & 23.0\cc{r8} & 1.3 \\ \midrule
\multirow{8}{*}{\textbf{\begin{tabular}[c]{@{}l@{}}deep-\\ lstm\end{tabular}}} & \textbf{ar-en} & 21.2\cc{r8} & 25.7\cc{r3} & 27.2\cc{r1} & 27.1\cc{r2} & 25.6\cc{r4} & 24.8\cc{r6} & 25.1\cc{r5} & 22.9\cc{r7} & 4.3 \\
 & \textbf{cs-en} & 19.8\cc{r6} & 22.0\cc{r1} & 18.5\cc{r7} & 21.1\cc{r3} & 20.9\cc{r4} & 21.2\cc{r2} & 20.3\cc{r5} & 15.8\cc{r8} & 6.2 \\
 & \textbf{de-en} & 25.7\cc{r1} & 25.2\cc{r2} & 24.9\cc{r3} & 24.1\cc{r5} & 24.5\cc{r4} & 23.5\cc{r6} & 23.5\cc{r6} & 23.1\cc{r8} & 2.6 \\
 & \textbf{fr-en} & 25.6\cc{r5} & 26.8\cc{r3} & 27.1\cc{r1} & 26.0\cc{r4} & 26.9\cc{r2} & 25.6\cc{r5} & 17.9\cc{r8} & 22.8\cc{r7} & 9.2 \\ \cmidrule(l){2-11} 
 & \textbf{en-ar} & 10.9\cc{r1} & 10.2\cc{r3} & 10.3\cc{r2} & 7.5\cc{r7} & 9.5\cc{r4} & 9.4\cc{r5} & 7.2\cc{r8} & 8.0\cc{r6} & 3.7 \\
 & \textbf{en-cs} & 13.7\cc{r4} & 14.6\cc{r2} & 15.3\cc{r1} & 14.6\cc{r2} & 12.2\cc{r7} & 12.6\cc{r5} & 11.9\cc{r8} & 12.6\cc{r5} & 3.4 \\
 & \textbf{en-de} & 22.4\cc{r7} & 24.9\cc{r1} & 23.6\cc{r5} & 23.9\cc{r4} & 22.4\cc{r7} & 24.0\cc{r3} & 24.3\cc{r2} & 23.4\cc{r6} & 2.5 \\
 & \textbf{en-fr} & 23.1\cc{r2} & 22.9\cc{r4} & 23.5\cc{r1} & 23.1\cc{r2} & 22.2\cc{r5} & 22.0\cc{r6} & 18.0\cc{r8} & 20.0\cc{r7} & 5.5 \\ \bottomrule
\end{tabular}}
\caption{BLEU score for LSTM architectures with multiple BPE configurations. Each score is color-coded by its rank among scores from different BPE configurations in the same row. $\delta$ is the difference between the best and worst BLEU score of each row. \label{tab:lstm}}
\end{table*}

\begin{table*}[t]
\centering
\scalebox{0.9}{\begin{tabular}{@{}llrrrrrrr@{}}
\toprule
 &  & \multicolumn{1}{c}{\textbf{Char}} & \multicolumn{3}{c}{\textbf{Separate BPE}} & \multicolumn{3}{c}{\textbf{Joint BPE}} \\ \cmidrule(l){4-9} 
 &  & \multicolumn{1}{l}{\textbf{}} & \multicolumn{1}{l}{\textbf{2k}} & \multicolumn{1}{l}{\textbf{8k}} & \multicolumn{1}{l}{\textbf{32k}} & \multicolumn{1}{l}{\textbf{2k}} & \multicolumn{1}{l}{\textbf{8k}} & \multicolumn{1}{l}{\textbf{32k}} \\ \midrule
\multirow{2}{*}{\textbf{ar-en}} & \textbf{src} & 0.49k & 2.48k & 8.47k & 32.36k & 2.46k & 7.98k & 26.11k \\
 & \textbf{tgt} & 0.24k & 2.23k & 8.17k & 30.45k & 1.27k & 4.06k & 13.45k \\ \midrule
\multirow{2}{*}{\textbf{fr-en}} & \textbf{src} & 0.30k & 2.30k & 8.26k & 31.23k & 2.18k & 7.14k & 24.48k \\
 & \textbf{tgt} & 0.23k & 2.22k & 8.16k & 30.40k & 1.94k & 6.10k & 20.45k \\ \bottomrule
\end{tabular}}
\caption{Vocabulary size after applying separate and joint BPE for ar-en and fr-en language pair. \label{tab:separate-joint-vocab}}
\end{table*}

\begin{table}[t]
\resizebox{\columnwidth}{!}{\begin{tabular}{@{}llrrrr@{}}
\toprule
\multicolumn{2}{l}{} & \multicolumn{1}{l}{\textbf{\begin{tabular}[c]{@{}l@{}}Best\\ Sep.\end{tabular}}} & \multicolumn{1}{l}{\textbf{\begin{tabular}[c]{@{}l@{}}Best\\ Joint\end{tabular}}} & \multicolumn{1}{l}{\textbf{\begin{tabular}[c]{@{}l@{}}Worst\\ Sep.\end{tabular}}} & \multicolumn{1}{l}{\textbf{\begin{tabular}[c]{@{}l@{}}Worst\\ Joint\end{tabular}}} \\ \midrule
\multirow{8}{*}{\textbf{\begin{tabular}[c]{@{}l@{}}tiny-\\ lstm\end{tabular}}} & \textbf{ar-en} & 24.3 & 24.2 & 20.6 & 20.6 \\
 & \textbf{cs-en} & 20.2 & 20.7 & 17.8 & 17.8 \\
 & \textbf{de-en} & 23.3 & 23.6 & 21.1 & 21.1 \\
 & \textbf{fr-en} & 25.0 & 25.4 & 21.8 & 21.8 \\ \cmidrule(l){2-6} 
 & \textbf{en-ar} & 9.1 & 9.3 & 8.3 & 8.5 \\
 & \textbf{en-cs} & 15.2 & 14.4 & 11.5 & 11.5 \\
 & \textbf{en-de} & 21.8 & 21.9 & 18.2 & 18.2 \\
 & \textbf{en-fr} & 21.1 & 21.8 & 19.9 & 19.9 \\ \midrule
\multirow{8}{*}{\textbf{\begin{tabular}[c]{@{}l@{}}deep-\\ transformer\end{tabular}}} & \textbf{ar-en} & 31.0 & 30.8 & 26.8 & 27.5 \\
 & \textbf{cs-en} & 24.6 & 24.6 & 19.0 & 20.6 \\
 & \textbf{de-en} & 28.1 & 28.6 & 24.8 & 25.2 \\
 & \textbf{fr-en} & 28.8 & 29.8 & 27.3 & 26.6 \\ \cmidrule(l){2-6} 
 & \textbf{en-ar} & 12.0 & 13.0 & 9.6 & 10.6 \\
 & \textbf{en-cs} & 17.3 & 17.3 & 13.0 & 13.8 \\
 & \textbf{en-de} & 27.3 & 27.4 & 23.8 & 23.9 \\
 & \textbf{en-fr} & 24.0 & 25.6 & 22.5 & 22.8 \\ \bottomrule
\end{tabular}}
\caption{Best and worst BLEU score with \texttt{tiny-lstm} and \texttt{deep-transformer} for joint and separate BPE models. \label{tab:sep-bpe}}
\end{table}

Table~\ref{tab:lstm} shows the BLEU score for LSTM-based architectures trained with BPE merge operations ranging from 0 to 32k.
Among the three tables, the \texttt{shallow-lstm} architecture has the minimal variation with regard to different merge operation choices.
For \texttt{tiny-lstm}, we observe a drastic performance drop between BPE merge operations 0/500 or 500/1k.
But aside from these two settings, the variation is of similar scale to \texttt{shallow-lstm}.
For \texttt{deep-lstm}, the variation is even larger than the Transformer architectures, and compared to \texttt{tiny-lstm} and \texttt{shallow-lstm}, the optimal BPE configuration shifts to BPE sizes on the smaller end.
However, we have also noticed that the overall absolute BLEU score of \texttt{deep-lstm} is lower than \texttt{shallow-lstm} despite more parameter is being used.
We conjecture that the larger variation and lower BLEU score from the \texttt{deep-lstm} experiments is largely due to the overfitting effect on the small training data. 
Despite this effect, moving from tiny to deep model, we observe a trend that deeper models tends to make use of smaller BPE size better.
In general, we conclude that unlike Transformer architecture, there is no typical optimal BPE configuration setting for the LSTM architecture. Because of this noisiness, we urge that future work using LSTM-based baselines tune their BPE configuration in a wider range on a development set to the extent possible, in order to ensure reasonable comparison.

\subsection{Analysis 2: Joint vs Separate BPE}

Another question that is not extensively explored in the existing literature is whether \textit{joint BPE} is the definitive better approach to apply BPE.
The alternative way, referred to here as separate BPE, is to build separate models for source and target side of the parallel corpus.
\newcite{DBLP:conf/acl/SennrichHB16a} conducted experiments with both joint and separate BPE, but these experiments were conducted with different BPE size, and not much analysis was conducted on the separate BPE model.
\newcite{DBLP:conf/wmt/HuckRF17} is the only other work we are aware of that used with separate BPE models for their study.
It was mentioned that their joint BPE vocabulary of 59500 yielded a German vocabulary twice as large as English, which is an undesirable characteristic for their study.

Before comparing the system performance, we would like to systematically understand how the resulting vocabulary is different when jointly and separately applying BPE.
Table~\ref{tab:separate-joint-vocab} shows the two most typical cases for this comparison, namely the Arabic-English language pair and the French-English language pair.
The reason these two language pairs are typical is that for Arabic-English, the scripts of the two languages are completely different, while the French and English scripts only have minor difference.
It could be seen that for Arabic-English language pair, the Arabic vocabulary size is always roughly twice the size of the English vocabulary.
Upon closer examination, we see that roughly half of the Arabic vocabulary is consisted of English words and subwords, scattering over around 2\% of the lines in the Arabic side of the training corpus.\footnote{These English tokens are generally English names, URLs or other untranslated concepts or acronyms.}
Hence, for most sentence pairs in the training data, the \emph{effective} Arabic and English vocabulary under joint BPE model is still roughly the same size. 
On the other hand, because of extensive subword vocabulary sharing, at lower BPE size, the vocabulary size for French and English is always roughly the same as the number of BPE merge operations regardless of separate or joint BPE.
However, this equality starts to diverge as more BPE merge operations are conducted, because the vocabulary difference between French and English starts to play out in this scenario.
Unlike Arabic-English, it is hard to predict what is the resulting BPE size from the number of merge operations used, because it is hard to know how many resulting subwords will be shared between the two languages.

Table~\ref{tab:sep-bpe} shows our experimental results with separate/joint BPE and our base architectures.\footnote{We only run experiments on 2k, 8k and 32k to save computation time.} With the configurations we explore, the difference between the best separate/joint BPE performance seems minimal.
On the other hand, while the worst BPE configuration remains the same for separate BPE models, we see even worse performance for Transformer at 32k separate BPE most of the time.
We think this is a continuation of the trend observed in our main results, as the vocabulary size tends to be even larger than joint BPE when applying separate BPE models.

Given the negligible difference in model performance, we think it is not necessary to sweep BPE merge operations for both joint and separate settings.
It is sufficient to focus on the setting that makes the most sense for the task at hand, and focus on hyper parameter search within that setting.


\subsection{Analysis 3: Languages}

\begin{table}[]
\centering
\resizebox{\columnwidth}{!}{\begin{tabular}{@{}lrrrllll@{}}
\toprule
 & \multicolumn{1}{l}{\textbf{0.5k}} & \multicolumn{1}{l}{\textbf{32k}} & \multicolumn{1}{l}{\textbf{$\delta$}} &  & \textbf{0.5k} & \textbf{32k} & \textbf{$\delta$} \\ \midrule
\textbf{pt-en} & 36.3 & 34.7 & 1.6 & \textbf{en-pt} & 38.5 & 35.6 & 2.9 \\
\textbf{he-en} & 31.1 & 28.6 & 2.5 & \textbf{en-he} & 26.2 & 22.9 & 3.3 \\
\textbf{tr-en} & 20.9 & 17.8 & 3.1 & \textbf{en-tr} & 13.0 & 9.8 & 3.2 \\
\textbf{ru-en} & 19.9 & 18.0 & 1.9 & \textbf{en-ru} & 19.1 & 16.6 & 2.5 \\
\textbf{pl-en} & 19.3 & 16.7 & 2.6 & \textbf{en-pl} & 16.7 & 13.4 & 3.3 \\
\textbf{hu-en} & 20.8 & 16.8 & 4.0 & \textbf{en-hu} & 16.0 & 12.6 & 3.4 \\ \bottomrule
\end{tabular}}
\caption{BLEU score for the 6 extra language pairs in multilingual-TED dataset with \texttt{deep-transformer} architecture. \label{tab:extra-lang}}
\end{table}

\begin{table}[]
\centering
\scalebox{0.9}{\begin{tabular}{@{}lrrr@{}}
\toprule
 & \multicolumn{1}{l}{\textbf{coef.}} & \multicolumn{1}{l}{\textbf{std. error}} & \multicolumn{1}{l}{\textbf{$p$-value}} \\ \midrule
$f_1$ & 0.575 & 1.345 & 0.677 \\
$f_2$ & -0.460 & 1.345 & 0.738 \\
$f_3$ & -1.998 & 1.983 & 0.333 \\
$f_4$ & 0.304 & 0.360 & 0.415 \\
$f_5$ & 1.060 & 0.639 & 0.123 \\
$f_6$ & 1.169 & 0.516 & 0.043 \\
$f_7$ & 0.913 & 0.314 & 0.013 \\
$f_8$ & 0.340 & 0.367 & 0.373 \\
$f_9$ & 1.280 & 0.755 & 0.116 \\ \bottomrule
\end{tabular}}
\caption{Coefficient from regression analysis and their corresponding standard error and $p$-values. $f_1$ and $f_2$ are source and target type/token ratio, respectively. $f_3$ is alignment ratio. $f_4$--$f_6$ are binary features for source-side morphological type (fusional, introflexive and agglutinative) and $f_7$--$f_9$ are the same for target. \label{tab:reg}}
\end{table}

We are interested in what properties of the language have the most impact on the variance of BLEU score with regard to different BPE configurations.
For our main experiments, we can already see a pretty consistent trend that for \texttt{deep-transformer} architecture, 0.5k and 32k merge operations always roughly correspond to the best and worst BPE configurations, respectively.
To add more data points, we assume 0.5k and 32k are always the best and the worst configurations and build systems with these two configurations with both translation directions of 6 more languages pairs, namely, translating of English into and out of  Brazilian Portuguese (pt), Hebrew (he), Russian (ru), Turkish (tr), Polish (pl) and Hungarian (hu).
Table~\ref{tab:extra-lang} shows the result with these 6 language pairs.
We note that our observation for the 4 language pairs generalize well for the extra 6 language pairs, and we observe a similar magnitude of performance drop as the other language pairs moving from 0.5k to 32k.

To acquire insights for the aforementioned problem, we conduct a linear regression analysis using the linguistic features of the the 10 language pairs as independent variables and BLEU score difference between 0.5k and 32k merge operation settings as the dependent variable.\footnote{Note that for language pairs in our main results, these may not necessarily the best or the worst system. But the readers shall see that the difference is pretty minimal.}
The linguistic features of our interest are described as follows:
\begin{itemize} \itemsep -2pt
\item \textbf{Type/Token Ratio}: Taken from \newcite{bentz2016comparison}  this is the ratio between number of token types and the number of tokens in the training corpus, ranging $[0, 1]$. These are computed separately for source and target language and denoted as $f_1$ and $f_2$ respectively.
\item \textbf{Alignment Ratio}: Also taken from \newcite{bentz2016comparison}, this is the relative difference between the number of many-to-one alignments and one-to-many alignments in the training corpus, ranging $[-1, 1]$. We follow the same alignment setting as in \newcite{DBLP:journals/corr/abs-1809-02223}. This is computed together for each parallel training corpus and denoted as $f_3$.
\item \textbf{Morphological Type}: We then use a set of binary features to indicate if a language exhibits a certain morphological patterns.
We take morphological features from \newcite{gerz-etal-2018-relation}, where for each language a morphological type from the following categories was assigned: \textit{Isolating}, \textit{Fusional}, \textit{Introflexive} and \textit{Agglutinative}.
None of the languages we use exhibit \textit{Isolating} morphology which leaves us with $6$ binary features. 
The features $f_4,f_5$ and $f_6$ indicates the presence (or absence) of \textit{fusional}, \textit{introflexive} and \textit{agglutinative} morphological patterns respectively for the source language and $f_7,f_8,f_9$ indicate the same for the target side.
\end{itemize}

\begin{figure*}[h]
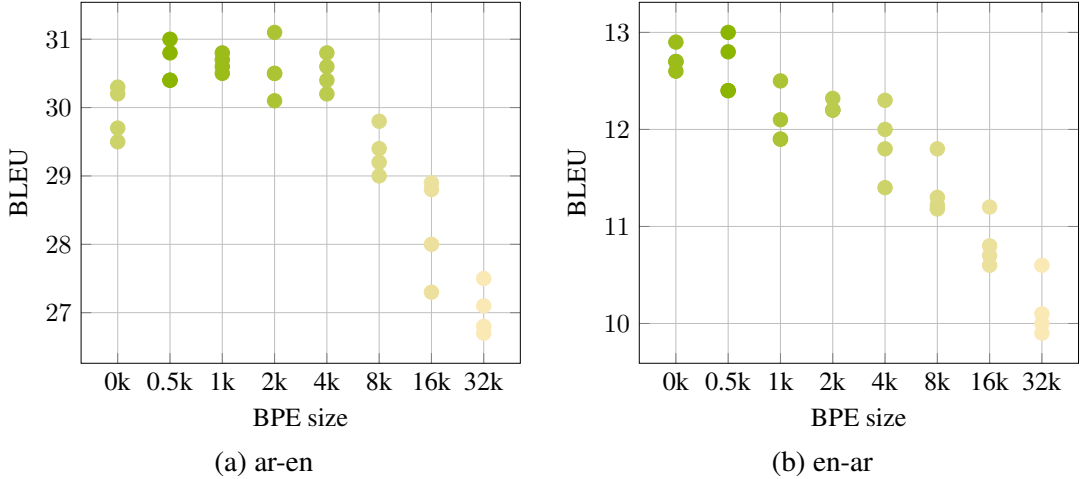

\centering
\begin{tabular}{cc}
\scalebox{0.9}{\import{}{figs/aren_variance2.plot}} &
\scalebox{0.9}{\import{}{figs/enar_variance2.plot}} \\
(a) ar-en & (b) en-ar \\
\end{tabular}
\caption{Scatter plots for the variance analysis of \texttt{deep-transformer} system. Each dot in the plot represents the BLEU score for one random restart, while the color code follows the result ranking of its corresponding system configuration in Table \ref{tab:transformer}. \label{fig:variance}}
\end{figure*}

The $9$ features are re-normalized to the $[0, 1]$ region with the min-max normalization.
Our linear regression analysis is conducted with Ordinary Least Squares (OLS) model in the Python statsmodels\footnote{\tt\scriptsize https://pypi.org/project/statsmodels/0.9.0/} package.

Table~\ref{tab:reg} shows the regression result.
Surprisingly, we don't see any strong correlation between the type/token ratio, alignment ratio and the variance in BPE.
On the other hand, the regression points out that having agglutinative language on the source side and fusional language on the target side increases such variance.
While we have seen significant BPE variances for all the experiments with Transformer, we think future work should be especially cautious with systems that translate out of agglutinative language and into fusional language (note that English is classified as fusional language in this regime).

\subsection{Analysis 4: Variance with Random Seeds}

Since our experiments are under low-resource settings, it is important to examine whether the trends we observe above are due to different system configurations or mostly variance of random seeds.
As it is expensive to re-run all the systems multiple times, we only conduct such analysis on the \texttt{deep-transformer} architecture and ar-en and en-ar language pairs.
We choose to focus on Transformer architecture because we observe more consistent trend for Transformer than LSTM.
Hence, it is more interesting to see how well it holds against the randomness in training.
To conduct such analysis, we run each system configuration for three more times with different random seeds resulting in four points for each system configuration.

Figure~\ref{fig:variance} shows the scatter plots of BLEU scores for each random restart under each system configuration.
Ideally, the BLEU scores from multiple random restarts of the system configurations should preserve the same ranking as the results in Table~\ref{tab:transformer}.
It can be seen that, the results from the top-3 BPE configurations are often clustered together (indicating low variance) and the rankings of the other configurations are preserved pretty well.
Specifically, even best instances among multiple random restarts with 16k and 32k BPE merge operations fall pretty far from those with top configurations, further verifying our previous observations on the Transformer architecture.
\subsection{Analysis 5: High-Resource Setting}
\begin{table*}[]
\centering
\scalebox{0.9}{\begin{tabular}{@{}lrrrrrrrrr@{}}
\toprule
 & \multicolumn{1}{l}{\textbf{0}} & \multicolumn{1}{l}{\textbf{0.5k}} & \multicolumn{1}{l}{\textbf{1k}} & \multicolumn{1}{l}{\textbf{2k}} & \multicolumn{1}{l}{\textbf{4k}} & \multicolumn{1}{l}{\textbf{8k}} & \multicolumn{1}{l}{\textbf{16k}} & \multicolumn{1}{l}{\textbf{32k}} & \multicolumn{1}{l}{\textbf{$\delta$}} \\ \midrule
\textbf{ru-en} & 29.3\cc{r8} & 30.4\cc{r5} & 30.0\cc{r7} & 30.3\cc{r6} & 30.6\cc{r4} & 30.9\cc{r2} & 31.0\cc{r1} & 30.9\cc{r2} & 1.7 \\
\textbf{en-ru} & 28.0\cc{r8} & 29.1\cc{r6} & 29.1\cc{r6} & 29.5\cc{r4} & 29.5\cc{r4} & 29.8\cc{r3} & 30.0\cc{r1} & 30.0\cc{r1} & 2.0 \\ \bottomrule
\end{tabular}}
\caption{BLEU score for \texttt{deep-transformer} architecture under high-resource setting, with multiple BPE configurations. Each score is color-coded by its rank among scores from different BPE configurations in the same row. $\delta$ is the difference between the best and worst BLEU score of each row. \label{tab:high-resource}}
\end{table*}
While this paper focuses on low-resource settings, we conduct one set of experiments with a high-resource language pair to see if our results generalize to high-resource settings.
This experiment is conducted with all WMT 2017 Russian-English (ru-en) data except the UN dataset, which includes 2.61M sentence pairs in total.
We use the test sets from news translation shared task of WMT 2012-2016 as the development data and test on WMT 2017 test set.
Due to computation constraints, we only experiment with \texttt{deep-transformer} architecture.
All the other configurations are exactly the same as the low-resource experiments.

Table~\ref{tab:high-resource} summarizes the results.
First, notice that the overall variance of results under different BPE configurations is relatively smaller than the low-resource experiments, verifying our intuition that it is especially important to tune BPE size under low-resource settings.
Besides, the trend in this setting is also very different from what is shown in Table~\ref{tab:transformer}.
Specifically, the best results are often obtained with larger BPE sizes, which explains why these configurations were preferred by previous analysis.
It could hence be concluded that the analysis results in this paper should \emph{not} be generalized to high-source settings.
We leave comprehensive analysis with high-resource language pairs for future work.

\section{Conclusion}
We conduct a systematic exploration over various numbers of BPE merge operations to understand its interaction with system performance.
We conduct this investigation over $5$ different NMT architectures including encoder-decoder and Transformer, and $4$ language pairs in both translation directions.
We leave systematic study on the effect of BPE on high-resource settings and more language pairs, especially morphologically isolating languages, for future work.
Subword regularization could also be studied in this manner.

Based on the findings, we make the following recommendations for selecting BPE merge operations in the future: 
\begin{itemize} \itemsep -2pt
    \item For Transformer-based architectures, we recommend the sweep be concentrated in the $0-4$k range.
    \item For Shallow LSTM architectures, we find no typically optimal BPE merge operation and therefore urge future work to sweep over $0-32$k to the extent possible.
    \item We find no significant performance differences between joint BPE and seperate BPE and therefore recommend BPE sweep be conducted with either of these settings.
\end{itemize}
Furthermore, we strongly urge that the aforementioned checks be conducted when translating into fusional languages (such as English or French) or when translating from agglutinative languages (such as Turkish).

Our hope is that future work could take the experiments presented here to guide their choices regarding BPE and wordpiece configurations, and that readers of low-resource NMT papers call for appropriate skepticism when the BPE configuration for the experiments appears to be sub-optimal.

\section*{Acknowledgments}
This work is supported in part by the Office of the Director of National 
Intelligence, Intelligence Advanced Research Projects Activity (IARPA), 
via contract \#FA8650-17-C-9115. The views and conclusions contained 
herein are those of the authors and should not be interpreted as 
necessarily representing the official policies, either expressed or 
implied, of the sponsors.


\bibliography{mtsummit2019}
\bibliographystyle{mtsummit2019}

\end{document}